\documentclass{article}

\usepackage{microtype}
\usepackage{graphicx}
\usepackage{subfigure}
\usepackage{booktabs} %

\usepackage{hyperref}

\usepackage{booktabs}
\usepackage{colortbl}

\usepackage{graphicx}
\usepackage{float,wrapfig}

\usepackage{array}
\usepackage{adjustbox}
\usepackage{multirow}
\usepackage{colortbl}

\usepackage{caption}
\usepackage{color,xcolor}
\usepackage{epsfig}
\usepackage{bbm}
\usepackage{hhline}

\usepackage{amsmath,amsfonts,amsthm,amssymb}
\usepackage{mathtools}
\usepackage{bm}
\usepackage{nicefrac}
\usepackage{microtype}

\usepackage{changepage}
\usepackage{extramarks}
\usepackage{fancyhdr}
\usepackage{lastpage}
\usepackage{setspace}
\usepackage{soul}
\usepackage{xspace}

\usepackage{url}
\usepackage{algorithm, algorithmic}
\usepackage{todonotes} %

\usepackage{titlesec}

\DeclareMathOperator*{\argmin}{arg\,min}

\newcommand{\sect}[1]{Section~\ref{#1}}

\newcommand{\eqn}[1]{Equation~\ref{#1}}
\newcommand{\fig}[1]{Figure~\ref{#1}}
\newcommand{\tbl}[1]{Table~\ref{#1}}
\newcommand{\myparagraph}[1]{{\bf #1}\quad}

\DeclarePairedDelimiterX{\infdivx}[2]{(}{)}{%
  #1\;\delimsize|\delimsize|\;#2%
}

\newcommand{\model}{IREM\xspace}

\definecolor{MyDarkBlue}{rgb}{0,0.08,1}
\definecolor{MyDarkGreen}{rgb}{0.02,0.6,0.02}
\definecolor{MyDarkRed}{rgb}{0.8,0.02,0.02}
\definecolor{MyDarkOrange}{rgb}{0.40,0.2,0.02}
\definecolor{MyPurple}{RGB}{111,0,255}
\definecolor{MyRed}{rgb}{1.0,0.0,0.0}
\definecolor{MyGold}{rgb}{0.75,0.6,0.12}
\definecolor{MyDarkgray}{rgb}{0.66, 0.66, 0.66}

\newcommand{\myitem}{\vspace{-5pt}\item}

\newtheorem{theorem}{Remark}

\usepackage{amsmath,amsfonts,bm}

\def\eqref#1{equation~\ref{#1}}

\def\1{\bm{1}}

\def\ve{{\bm{e}}}

\def\vs{{\bm{s}}}

\def\vv{{\bm{v}}}

\def\vx{{\bm{x}}}
\def\vy{{\bm{y}}}
\def\vz{{\bm{z}}}

\DeclareMathAlphabet{\mathsfit}{\encodingdefault}{\sfdefault}{m}{sl}
\SetMathAlphabet{\mathsfit}{bold}{\encodingdefault}{\sfdefault}{bx}{n}

\usepackage[accepted]{icml2022}

\icmltitlerunning{Learning Iterative Reasoning through Energy Minimization}

\begin{document}

\twocolumn[
\icmltitle{Learning Iterative Reasoning through Energy Minimization}

\icmlsetsymbol{equal}{*}

\begin{icmlauthorlist}
\icmlauthor{Yilun Du}{mit}
\icmlauthor{Shuang Li}{mit}
\icmlauthor{Joshua Tenenbaum}{mit}
\icmlauthor{Igor Mordatch}{goo}
\end{icmlauthorlist}

\icmlaffiliation{mit}{MIT CSAIL}
\icmlaffiliation{goo}{Google Brain}

\icmlcorrespondingauthor{Yilun Du}{yilundu@mit.edu}

\icmlkeywords{Energy Based Models, Optimization, Reasoning, Algorithms, Dynamic Programming}

\vskip 0.3in
]

\printAffiliationsAndNotice{} %

\begin{abstract}
Deep learning has excelled on complex pattern recognition tasks such as image classification and object recognition.
However, it struggles with tasks requiring nontrivial reasoning, such as algorithmic computation. 
Humans are able to solve such tasks through iterative reasoning -- spending more time thinking about harder tasks.
Most existing neural networks, however, exhibit a fixed computational budget controlled by the neural network architecture, preventing additional computational processing on harder tasks. 
In this work, we present a new framework for iterative reasoning with neural networks.
We train a neural network to parameterize an energy landscape over all outputs, and implement each step of the iterative reasoning as an energy minimization step to find a minimal energy solution.
By formulating reasoning as an energy minimization problem, for harder problems that lead to more complex energy landscapes,  we may then adjust our underlying computational budget by running a more complex optimization procedure.
We empirically illustrate that our iterative reasoning approach can solve more accurate and generalizable algorithmic reasoning tasks in both graph and continuous domains. 
Finally, we illustrate that our approach can recursively solve algorithmic problems requiring nested reasoning. Code and additional information is available at  \href{https://energy-based-model.github.io/iterative-reasoning-as-energy-minimization/}{https://energy-based-model.github.io/iterative-reasoning-as-energy-minimization/}.
\end{abstract}
\section{Introduction}

Human thinking is often characterized in terms of mechanisms for two distinct modes of cognitive processing: {\sc System 1} mechanisms for fast, habitual, and associative processing, and {\sc System 2} mechanisms for slower, more deliberate and controlled, symbolic reasoning \citep{kahneman2011thinking}. Neural networks have excelled at habitual {\sc System 1}-style processing in familiar environments and task contexts, such as mapping images of familiar objects to their semantic classes, or familiar locations to the corresponding routes of movement. However, confronted with a novel environment or task demanding a more flexible response, humans can invoke controlled {\sc System 2} processes, often in the form of iteratively reasoning about relationships between observed entities that builds on past experiences primarily through shared abstractions and algorithms rather than direct re-use of concrete, habitual responses. %
Such flexible processing of novel inputs is difficult for neural networks, with even large pretrained language models \citep{radford2019language} that have proven effective in many zero-shot generalization contexts failing to extrapolate the simplest algorithmic operations, such as addition, to more complex inputs.

\begin{figure}[t]
\centering

\includegraphics[width=0.95\linewidth]{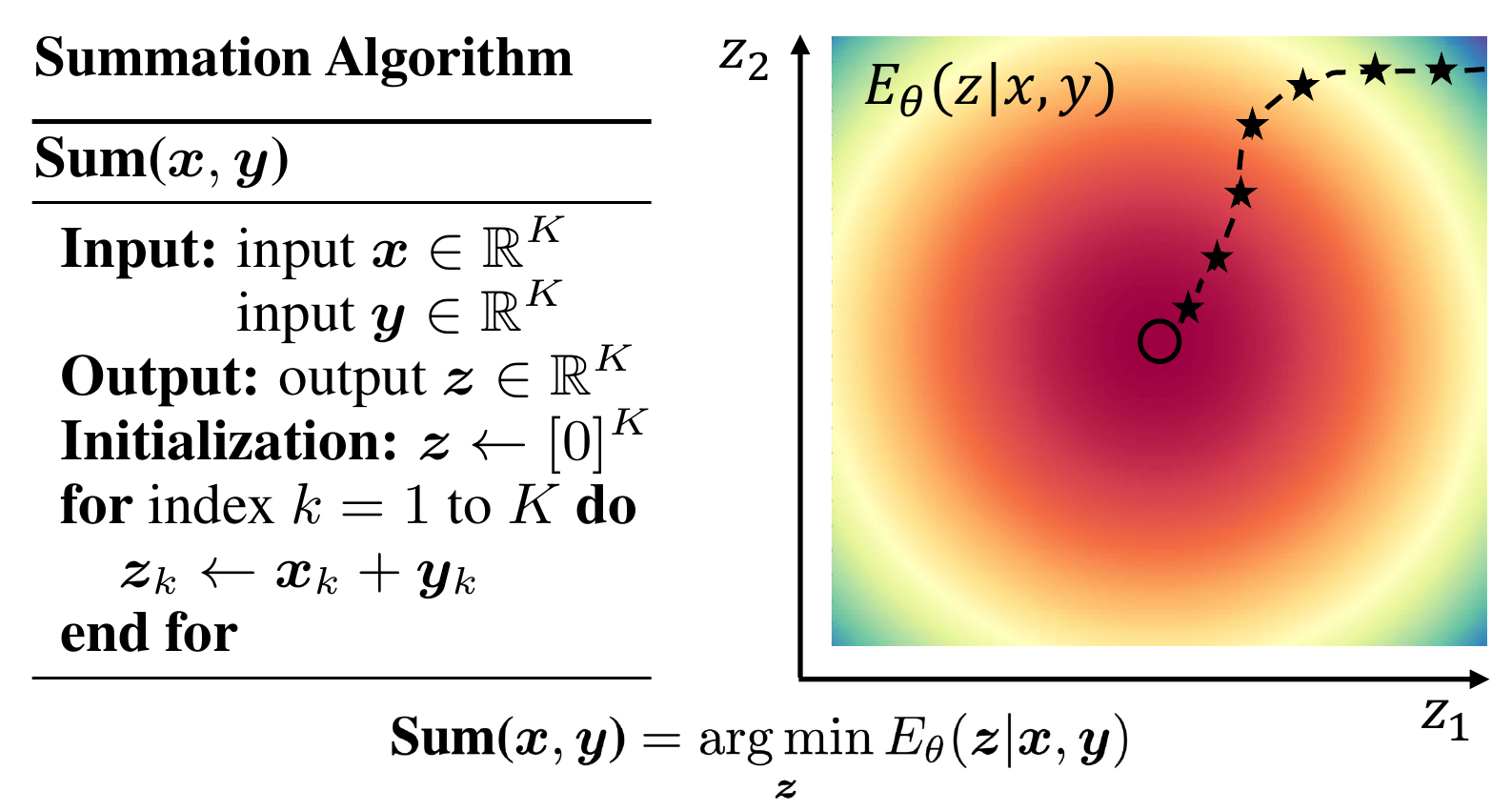}
\vspace{-10pt}
\caption{\small \textbf{Reasoning as Energy Minimization} -- \model formulates an algorithmic reasoning problem with inputs $\vx, \vy$ and output $\vz$, such as summation (\textbf{left}), as an iterative energy minimization problem over a learned energy function $E_\theta(\vz|\vx, \vy)$ which parameterizes an energy landscape over all possible outputs $\vz$ conditioned on inputs $\vx, \vy$ (\textbf{right}). We visualize the first two dimensions $\vz_1$ and $\vz_2$. Each iteration of energy minimization is shown by a star. }
\label{fig:teaser}
\vspace{-15pt}
\end{figure}

Iterative reasoning, the ability to repeatedly apply an underlying computation to the outputs of previous reasoning steps, is a crucial component of scalable {\sc System 2}-style processing. We may take abstractions learned from simpler variants of a problem, and iteratively apply them to solve harder variants of potentially unbounded complexity. As an example, having learned how to compute shortest paths on small graphs, we might generalize and apply our algorithmic intuition sequentially to substantially larger graph problems at test time through iterative application of learned computations. Such iterative processing enables humans to solve more challenging tasks, such as question answering, arithmetic calculation, or proof writing, even in novel or unfamiliar domains. %

We present a new neural approach towards iterative reasoning, which we formulate as an energy minimization process on a learned energy landscape (\fig{fig:teaser}). By representing individual steps of reasoning as an optimization process, we may iteratively reason for longer on harder problems by running additional steps of optimization on the induced energy landscapes. Simultaneously, by monitoring the geometric energy landscape surrounding an optimized solution, we may automatically determine the completion of the algorithmic computation (by checking the presence of a local energy minimum). We refer to our underlying reasoning framework as Iterative Reasoning as Energy Minimization (\model).

To evaluate and benchmark the effectiveness of iterative reasoning using \model, we propose and construct a suite of different algorithmic reasoning tasks on both graph and continuous domains. Effective algorithmic reasoning requires repetitive application of underlying algorithmic computations, dependent on problem complexity, and thus serves as a natural benchmark for iterative reasoning. We compare \model with past works on our algorithmic benchmark for iterative reasoning and find that \model outperforms prior works in performance and generalization.

Our contributions in this paper are threefold. First, we present \model, a new framework for iterative reasoning and analyze why it is beneficial to use such a framework for reasoning. Second, we present a benchmark for iterative algorithmic reasoning, both on graphs and continuous vector inputs, and show that \model significantly outperforms prior approaches in both performance and generalization. Finally, we show how our approach can recursively solve algorithmic computation requiring nested reasoning. Our results point to \model as a promising new approach towards iterative reasoning.

\section{Related Work}

\myparagraph{Iterative Reasoning} A variety of recent works have explored the integration of iterative reasoning into neural networks. One branch of works implements iterative reasoning by constructing neural programmatic operations \citep{graves2014neural, reed2015neural, banino2021pondernet} which are repeatedly executed till halting. Another branch of work implements iterative reasoning through recurrent computation \citep{graves2016adaptive, bolukbasi2017adaptive, chung2016hierarchical, schwarzschild2021can}. A key challenge with both types of approaches lies in the halting time of computation. Existing approaches learn halting policies through reinforcement learning \citep{chen2020learning, chung2016hierarchical}, heuristic policies \citep{bolukbasi2017adaptive}, or variational inference \citep{graves2016adaptive, banino2021pondernet}. Such approaches are unstable in nature \citep{banino2021pondernet}, and many of them require manual hyper-parameter specification. We present an orthogonal approach towards implementing iterative reasoning as an energy minimization procedure on a learned energy landscape. By determining when a local energy minimum has been found, our approach provides a natural mechanism for terminating computation.

\myparagraph{Algorithmic Reasoning with External Memory} Several approaches towards iterative reasoning utilize an external memory scratchpad for algorithmic computation. Such a scratchpad enables models to store intermediate algorithmic computations, and thus boosts the underlying performance of the algorithm \citep{graves2014neural, reed2015neural, cai2017making, kaiser2015neural}. In the Appendix B, we illustrate a manner through which we may utilize an external memory with \model to improve underlying further improve reasoning performance.

\myparagraph{Optimization Based Computational Blocks} Prior works have explored optimization as a computation block to solve different tasks. In \citep{brockett1991dynamical}, a dynamical system is constructed that can solve various algorithmic tasks. Optimization has since been used as an intermediate neural network computation block for quadratic programs \citep{amos2017optnet, donti2017task} and submodular programs \citep{djolonga2017differentiable, wilder2019melding} for flexible neural networks. Most similar to our work, \citet{bai2019deep} utilizes equilibrium energy minimization as an intermediate computation block for memory-efficient neural networks. In contrast, we explore how direct optimization over a learned energy landscape can enable  generalizable iterative reasoning. Concurrent to our work, \citep{rubanova2021constraint} utilizes energy minimization to simulate physical dynamics.

\myparagraph{Energy-Based Models} Our work is related to works in Energy-Based Models (EBMs) \citep{lecun2006tutorial}. Most recent works using EBMs have focused on learning probabilistic models over data \citep{du2019implicit, nijkamp2019anatomy, grathwohl2020cutting, du2020compositional, du2021improved, arbel2020ebm,li2020energy, xiao2020vaebm}. Instead of using EBMs as a probabilistic model, we use EBMs to define an energy landscape for solving iterative reasoning problems.

\myparagraph{Learning Optimizers} Our work utilizes backpropagation through intermediate optimization steps to train our energy function. Prior work has explored a similar idea of backpropagation through optimization to learn meta optimizers \citep{andrychowicz2016learning, ravi2016optimization, bengio1995optimization, schmidhuber1992learning, hochreiter2001learning, li2016learning} which enable more sample efficient and faster training of neural networks. In this work, we utilize this approach to learn an energy function so that a number of optimization steps on the energy function results in a solution to the algorithmic reasoning problem.

\section{Learning Iterative Reasoning through Energy Optimization}

Let $\mathcal{D} = \{X, Y\}$ be a dataset of algorithmic reasoning problems consisting of inputs $\vx \in \mathbb{R}^O$ and corresponding solutions $\vy \in \mathbb{R}^M$. Our goal is to learn a neural network operator $\text{Alg}_\theta(\cdot)$ which can generalize $\text{Alg}_\theta(\vx')$ to a test input $\vx' \in \mathbb{R}^{O'}$  , where $\vx'$ can be significantly larger and more challenging than the training data $\vx \in X$. 
We present our approach, \model, to adaptive algorithmic computation in \sect{sect:energy_min}, and discuss how to learn such an energy landscape in \sect{sect:energy_consistent}. We further provide analysis on why such a framework is favorable in \sect{sect:energy_analysis}.

\subsection{Iterative Reasoning as Energy Minimization}
\label{sect:energy_min}

Iterative reasoning is typically formulated as the repeated application of a neural network operator $f_\theta(\vx, \vy): \mathbb{R}^O \times \mathbb{R}^M \rightarrow \mathbb{R}^M $, to generate partial solutions $\vy^t$
\begin{equation}
   \vy^t = f_\theta(\vx, \vy^{t-1}),
   \label{eqn:iterative}
\end{equation}
where the final prediction $\vy = \vy^T$ is output after $T$ successive applications of $f_\theta$. The termination of iterative computation is often difficult to specify, with methods typically utilizing a halting policy that is difficult to train \citep{graves2016adaptive, chung2016hierarchical}. 

In this work, we propose an alternative approach towards iterative computation with a natural termination criteria, \model, which represents iterative reasoning as an optimization process over an Energy-Based Model (EBM) $E_\theta(\vx, \vy): \mathbb{R}^O \times \mathbb{R}^M \rightarrow \mathbb{R}$:
\begin{equation}
    \vy = \argmin_{\vy} E_\theta(\vx, \vy).
\end{equation}
An individual reasoning step is then represented as
\begin{equation}
   \vy^t = \vy^{t-1} - \lambda \nabla_{\vy} E_\theta(\vx, \vy^{t-1}),
   \label{eqn:energy_iterative}
\end{equation}
where $\lambda$ is the step size of each gradient descent step. We repeat the iterative computation in \eqn{eqn:energy_iterative} until $\vy^t$ is a local minima of the energy landscape, $E_\theta(\vy^t) = E_\theta(\vy^{t-1})$, where additional steps of optimization do not change the energy value of $\vy^t$. Finding a local minimum of the energy landscape thus provides us with a natural mechanism to halt iterative computation.

\subsection{Learning Energy Landscapes for Reasoning}
\label{sect:energy_consistent}

We next discuss how to learn \model. Given an input problem $\vx_i$ with a unique solution $\vy_i$, the simplest method to learn $E_\theta(\vx, \vy)$ is to directly supervise the minimal energy state $\argmin_{\vy}  E_{\theta}(\vx_i, \vy)$ with $\vy_i$ through regression %
\begin{equation}
      \mathcal{L}_{\text{MSE}}(\theta) = \| \argmin_{\vy}  E_{\theta}(\vx_i, \vy) - \vy_i \|^2.
\end{equation}
However, in practice, during training time, it is computationally expensive to compute $\argmin_{\vy} E_\theta(\vx, \vy)$, as the underlying energy landscape may be complex.

As a fast approximation to $\argmin_{\vy} E_\theta(\vx_i, \vy)$, we may approximate the $\argmin$ operation with respect to $\vy_i$ via $N$ steps of gradient optimization. An approximate optimum $\vy_i^N$ is obtained by:
\begin{equation}
    \vy_i^N = \vy_i^{N-1} - \lambda \nabla_{\vy} E_{\theta} (\vx_i, \vy_i^{N-1}).
    \label{eqn:opt_mult}
\end{equation}
In \eqn{eqn:opt_mult}, we initialize $\vy_i^0$ using uniform noise, with $\lambda$ denoting the step size for each gradient step, and sample $\vy_i^N$ corresponding to the result after $N$ steps of gradient descent. We then minimize the corresponding loss:
\begin{equation}
      \mathcal{L}_{\text{MSE}}(\theta) = \| \vy_i^N - \vy_i \|^2,
\end{equation}
where we may directly differentiate through the underlying optimization procedure \citep{finn2017model}. To alleviate the computational burden of computing second-order gradients across optimization steps, we empirically found that simply truncating back-propagation to the last optimization step maintained good performance at a significantly faster training speed as illustrated in \tbl{tbl:ablation}.

An underlying difficulty of utilizing $\vy_i^N$ as an approximation of $\argmin_{\vy} E_\theta(\vx_i, \vy)$ is that since only a small finite number of steps of gradient descent is applied, $\vy_i^N$ may be far from reaching $\argmin_{\vy} E_\theta(\vx_i, \vy)$. As a result, when running a larger number of iterative reasoning steps at test time to more precisely compute $\argmin_{\vy} E_\theta(\vx_i, \vy)$, our underlying solution may degrade.

To remedy this issue, we maintain a replay buffer of previously optimized samples $\vy_i^N$, and initialize $\vy_i^0$ either from previously optimized values $\vy_i^N$ or uniform noise. By initializing gradient descent optimization with previously optimized samples, we ensure that 
these samples are closer to $\argmin_{\vy} E_\theta(\vx_i, \vy)$.  A similar application of replay buffers has been utilized to train EBMs for consistent probability landscapes \citep{du2019implicit}.

We provide pseudocode for training \model in Algorithm \ref{alg:train} and executing algorithmic reasoning with \model in Algorithm \ref{alg:test}, where an energy minimum is determined at test time once the energy value of a solution does not change for certain iterations. A fixed learning rate, $\lambda = 100$, is used for training \model. At test time, $\lambda$ is empirically tuned so that  energy values decrease smoothly across iterations of optimization (harder problems require smaller $\lambda$ to optimize).  In Appendix \ref{sect:scratchpad}, we discuss how we may utilize optimization to further incorporate an external scratchpad into iterative reasoning using \model.  

As \model directly trains an energy landscape by optimizing samples to regress solutions, there is no guarantee that a smooth underlying energy landscape $E(\vx, \vy)$ is learned. Empirically, as seen in \fig{fig:teaser} and \fig{fig:energy_landscape}, we find that our objective does lead to consistent energy landscapes similar to past work \citep{du2021comet}.

\begin{algorithm}[t]
\small
\begin{algorithmic}
    \STATE \textbf{Input:} Data Dist $p_D(\vx, \vy)$, Replay Buffer $\mathcal{B}$, Step Size $\lambda$, Number of Steps $N$, EBM $E_\theta(\cdot)$, Uniform Distribution $U(-1, 1)$
    \STATE $\mathcal{B} \gets \varnothing$
    \WHILE{not converged}
    
    \STATE \emph{$\triangleright$ Sample data and candidate solutions from $p_d$ and replay buffer $\mathcal{B}$}
    \STATE $\vx_i, \vy_i \sim p_D$, $\tilde{\vy}_i^0 \sim \mathcal{U}(-1, 1)$
    \STATE $\vx_i^b, \vy_i^b, \tilde{\vy}_i^{b} \sim B$
    \STATE $\vx_i, \vy_i, \tilde{\vy}_i^0 \gets \vx_i \cup \vx_i^b, \vy_i \cup \vy_i^b, \tilde{\vy}_i^0 \cup \tilde{\vy}_i^b$
    \vspace{2mm}
    
    \STATE \emph{$\triangleright$ Generate low energy solutions through optimization:}
    \FOR{sample step $n = 1$ to $N$}
    \STATE $\tilde{\vy}_i^n \gets \tilde{\vy}_i^{n-1} -  \lambda \nabla_\vy E_\theta (\vx_i, \tilde{\vy}_i^{n-1})$
    \ENDFOR 
    \vspace{2mm}
    
     \STATE \emph{$\triangleright$ Optimize objective $\mathcal{L}_{\text{MSE}}$ wrt $\theta$:} 
    \STATE $\Delta \theta \gets \nabla_\theta \sum_{n=1}^N \|\tilde{\vy}^N_i - \vy_i\|^2 $
    \STATE Update $\theta$ based on $\Delta \theta$ using Adam optimizer 
    
    \vspace{2mm}
    \STATE \emph{$\triangleright$ Update replay buffer $\mathcal{B}$}
    \STATE $\mathcal{B} \gets \mathcal{B} \cup (\vx_i, \vy_i, \tilde{\vy}_i^{N})$
    \ENDWHILE

  \end{algorithmic}
 \caption{\model training algorithm}
 \label{alg:train}
 \end{algorithm}

\subsection{Analysis} 
\label{sect:energy_analysis}

In this section, we provide complexity-theoretic motivation for representing iterative reasoning as an energy minimization is advantageous. We consider two approaches to represent reasoning:

\myparagraph{Feedforward Computation.} We consider reasoning as a feedforward function $f$ of the form $f(\vx)$ where $f(\vx): \mathbb{R}^O \rightarrow \mathbb{R}^M$ maps a input $\vx$ to a predicted solution $\vy$. 

\myparagraph{Energy Minimization.} We next consider reasoning as a energy minimization problem, $\argmin_\vy E(\vx, \vy)$, where $E(\vx, \vy)$ is a function from $\mathbb{R}^{O} \times \mathbb{R}^M \rightarrow \mathbb{R}$, mapping a input $\vx$ and solution $\vy$ into an energy. This corresponds to the computation represented by \model.

We construct a class of algorithmic reasoning tasks that are easier to learn using an energy function $E(\vx, \vy)$ as opposed to the feedforward function $f(\vx)$.  Our result is based on the intuition that learning the energy function $E(\vx, \vy)$ corresponds to learning a solution verifier, which assigns minimal energy to the correct solution, and high energy to all the other solutions. In contrast, learning a function $f(\vx)$ corresponds to explicitly generating a solution. We rely on the well-known theorem in complexity theory that constructing a solution verifier is easier than a solution generator to show that learning energy minimization is easier than feedforward computation.

\begin{algorithm}[t]
\small
\begin{algorithmic}
    \STATE \textbf{Input:}  Data Dist $p_D(\vx)$, Step Size $\lambda$, Number of Steps $K$, EBM $E_\theta(\cdot)$, Uniform Distribution $U(-1, 1)$
    \STATE \emph{$\triangleright$ Sample input from $p_d$ and initialize candidate solution}
    \STATE $\vx_i \sim p_D$
    \STATE $\tilde{\vy}_i \sim \mathcal{U}(-1, 1)$
    \vspace{2mm}
    \WHILE{Not at Energy Minima}
    \STATE \emph{$\triangleright$ Optimize candidate solution $\tilde{\vy}_i$ with gradient descent:}
    \STATE $\tilde{\vy}_i \gets \tilde{\vy}_i -  \lambda \nabla_\vy E_\theta (\vx_i, \tilde{\vy}_i)$
    \ENDWHILE
    \vspace{2mm}
    \STATE \emph{$\triangleright$  Final predicted solution:}
    \STATE $\vy = \tilde{\vy}$
  \end{algorithmic}
 \caption{\model prediction algorithm}
 \label{alg:test}
 \end{algorithm}

\begin{table*}
\small\setlength{\tabcolsep}{5.5pt}
\centering
\scalebox{0.95}{
\begin{tabular}{llcc|llcc}
\toprule
      {\bf Task} & {\bf Method} & {\bf Same Size} & {\bf Larger Size} &  {\bf Task} & {\bf Method}  & {\bf Same Diff.} & {\bf Harder Diff.}\\
      \midrule
     \multirow{4}{*}{{\bf Edge}} & Feedforward & 0.3016 &  0.3124  & \multirow{5}{*}{{\bf Add}}  & Feedforward & 0.0448  & 0.7029  \\
      \multirow{4}{*}{{\bf Copy}}  & Recurrent &  0.3015  & 0.3113  &  & Recurrent & 0.3610 & 2.6133 \\
      & Programmatic & 0.3053 & 0.4409  & & Programmatic &  0.0111 & 0.3446    \\
      & Iterative Feedforward & 0.6163 & 0.6498  & & Iterative Feedforward & 0.0144 & 0.1577   \\
      & \model (Ours) & {\bf 0.0019} & {\bf 0.0019} & & \model (Ours) & \textbf{0.0003} & \textbf{0.0021}  \\
      \midrule
     \multirow{4}{*}{ {\bf Connected}} & Feedforward & 0.1796  &   0.3460  & \multirow{4}{*}{ {\bf Matrix}} & Feedforward & 0.0203 & 0.2720   \\
      \multirow{4}{*}{ {\bf Components}} & Recurrent &  0.1794 &  0.2766 & \multirow{4}{*}{ {\bf Completion}} & Recurrent & 0.0266 & 0.3285  \\
      & Programmatic & 0.2338 & 3.1381  &
       & Programmatic &  0.0203 & 0.2637  \\
      & Iterative Feedforward &  0.4908 & 1.2064  & & Iterative Feedforward &   0.0253 & 0.2102   \\
      & \model (Ours)  & \textbf{0.1424} & \textbf{0.2171} &  & \model (Ours)  & \textbf{0.0183} & \textbf{0.2074}\\
      \midrule
      \multirow{4}{*}{ {\bf Shortest}} & Feedforward & 0.1233 & 1.4089  & \multirow{4}{*}{ {\bf Matrix}} & Feedforward & 0.0112 & 0.2150   \\
      \multirow{4}{*}{ {\bf Path}} & Recurrent &   0.1259 & 0.1083   &  \multirow{4}{*}{ {\bf Inverse}} & Recurrent & 0.0109 &   0.2123 \\
      & Programmatic &   0.1375 & 0.1290 & & Programmatic &   0.0124  & 0.2209   \\
      & Iterative Feedforward &   0.4588 & 0.7688  & & Iterative Feedforward &  0.0270 & 0.5250   \\
      & \model (Ours) & \textbf{0.0274} & \textbf{0.0464} & & \model (Ours) & \textbf{0.0108} & \textbf{0.2083} \\
    \bottomrule
\end{tabular}
}
\caption{\textbf{Algorithmic Reasoning with \model --} \model is general framework for iterative reasoning which can learn algorithmic computation on both graph (\textbf{left}) and continuous (\textbf{right}) inputs. \model generalizes at test time to both larger (left) and harder (right) algorithmic problems through iterative computation. Error reported on each task using elementwise mean square error. Approaches on the left are trained on graphs with ten nodes and evaluated on larger graphs with fifteen nodes.  \model significantly outperforms comparisons.}
\label{tbl:tbl_algorithm}
\vspace{-10pt}
\end{table*}
To analyze $E(\vx, \vy)$, we consider solving the 3-SAT \citep{impagliazzo_paturi_1999} problem. We construct a energy function $E(\vx, \vy)$ to verify the 3-SAT formula. Given a 3-SAT formula $\phi$ with $D$ variables and $K$ clauses,  we construct an energy function to represent $\phi$ as $E(\vx, \vy) := \sum_{1 \leq k \leq K} e_k(\vx, \vy)$, where $\vx$ encodes the clauses in a 3-SAT problem, $\vy$ corresponds to boolean assignments to each variable, and $e_k$ verifies whether the boolean assignments in $\vy$ satisfies the $k^{\text{th}}$ clause. To encode a set of K clauses using $\vx$, we utilize an ordinal representation (e.g. $\vx_1 = [1, 2, 3]$ to represent a particular clause $(\vy_1 \land \vy_2 \land \vy_3)$). The energy function $e_i$ is then constructed by taking in clause $\vx_i$ and outputting 0 if the corresponding entries of $\vy$ satisfy the encoded clause and 1 otherwise (polynomial time to compute). We assume the Exponential Time Hypothesis (ETH) \citep{impagliazzo_paturi_1999}, which states that checking the satisfiability of a $3$-SAT formula takes time exponential in the sum of the number of variables and the number of clauses. 
\begin{theorem}
There exists a 3-SAT problem which may be encoded in an energy function $E(\vx, \vy)$ which can be evaluated at any input in time polynomial in the number of dimensions of $\vx$ but for which the computational complexity of encoding a feedforward solution $f(\vx)$ which may be evaluated at any input is (worse-case) exponential in the number of dimensions of $\vx$. 
\end{theorem}
\vspace{-15pt}
\begin{proof}
We encode the 3-SAT energy function $E(\vx, \vy)$ as defined above, which is evaluated in time polynomial in the number of dimensions of $\vx$. In contrast, ETH directly implies that constructing $f(\vx)$, which corresponds to solving the 3-SAT problem, is exponential in dimension of $\vx$.
\end{proof}
\vspace{-10pt}
Our remark shows that it is computationally advantageous to represent an energy function $E(\vx, \vy)$ as opposed to a feedforward decoder $f(\vx)$. In particular, learning neural networks $E_\theta(\vx, \vy)$ and $f_\theta(\vx)$ to approximate either $E(\vx, \vy)$ or $f(\vx)$, our remark implies that a larger network is necessary to represent the exponential computations of $f(\vx)$. Such a network $f_\theta(\vs)$ would require either exponentially deeper or wider layers, if there is no iterative computation. With iterative computation, we may parameterize $f_\theta(\vx)$ and $E_\theta(\vx, \vy)$ with a similar number of parameters, but then it would be necessary for $f_\theta(\vx)$ to be iteratively applied a exponential number of times. Direct recurrent backpropagation through such a number of iterative computations has been proven to be unstable to train. While in principle computing $\argmin_\vy E(\vx, \vy)$, would require a similar number of computations, we may train $E(\vx, \vy)$ with a simple inexact energy minimization procedure and run a more extensive minimization procedure at test time. We next analyze the computational complexity of representing $E(\vx, \vy)$ and $f(\vx)$ as a function of problem size.

\begin{theorem}
As the underlying number of variables in 3-SAT problem increases, we may construct $E(\vx, \vy)$ which can be evaluated at any input in time polynomial in the number of variables, but for which the computational complexity of encoding a feedforward solution $f(\vx)$ which may be evaluated at any input is (worse-case) exponential in the number of underlying variables. 
\end{theorem}
\vspace{-15pt}
\begin{proof}
Our constructed energy function $E(\vx, \vy)$ above may be evaluated in time polynomial in the number of input variables. In contrast, ETH implies that constructing $f(\vx)$, which corresponds to solving the 3-SAT problem, is exponential in the number of underlying variables.
\end{proof}
\vspace{-10pt}
Similar to our previous remark, our result implies that representing $f(\vx)$ requires exponentially more computation as the underlying problem size of a 3-SAT problem increases.  In particular, our remark has implications for {\bf generalization}. If we learn $f_\theta(\vx)$ and $E_\theta(\vx, \vy)$ on smaller problem instances of $\vx$, for our neural networks to generalize correctly to larger problems instances, the underlying computation executed by the neural network must increase polynomially for $E_\theta(\vx, \vy)$ and exponentially for $f_\theta(x)$. Existing architectures, such as transformers and graph networks, adaptively increase computation polynomially for larger inputs. In contrast, few architectures can adaptively increase computation exponentially for larger problem instances. To realize $f(\vx)$, we thus require iterative approaches which can execute exponentially longer on larger inputs. As seen in \fig{fig:steps_problem}, this is difficult for existing approaches. 
\section{Experiments}

\subsection{Experimental Setup}

We compare \model with feedforward and iterative baselines for reasoning. We discuss each approach in detail below. 

\myparagraph{Feedforward Computation.} First, we compare with (one-step) feedforward computation, where we train a neural network that directly outputs the values of solutions.

\myparagraph{Recurrent Network Computation.} Next, we compare our approach with methods utilizing a recurrent neural network to execute iterative computation. Recurrent architectures have been shown to successfully execute reasoning recently \citep{schwarzschild2021can}. We use a LSTM network to represent iterative computation.

\myparagraph{Learned Programmatic Computation.} We compare our method with past works which construct iterative computation through building programmatic structures with neural networks. We compare our method with the recent architecture and training objective of PonderNet \citep{banino2021pondernet}, which variationally learns both a halting probability and individual computation step networks.

\myparagraph{Iterative Feedforward Computation.} We further compare our approach with direct iterative application of a feedforward computation. We train the iterative feedforward computation using an iterative denoising objective \citep{sohl2015deep}. 

We scale network sizes to ensure that each individual baseline has roughly the same number of parameters. The architectural details of each model are provided in Appendix \ref{sect:model_architecture} of the paper. We utilize MLP neural networks for continuous algorithmic reasoning tasks and graph neural networks for graph algorithmic reasoning tasks. All iterative methods are trained with 5 steps of reasoning. We provide comparisons with each baseline on graphical algorithmic reasoning in \sect{sect:graph} and on continuous algorithmic reasoning in \sect{sect:continuous}. We provide additional benchmark comparisons of  \model on an existing iterative image denoising task in Appendix \ref{sect:image_denoise}.

\subsection{Graphical Algorithmic Reasoning}
\label{sect:graph}

\begin{figure}[t]
\includegraphics[width=\linewidth]{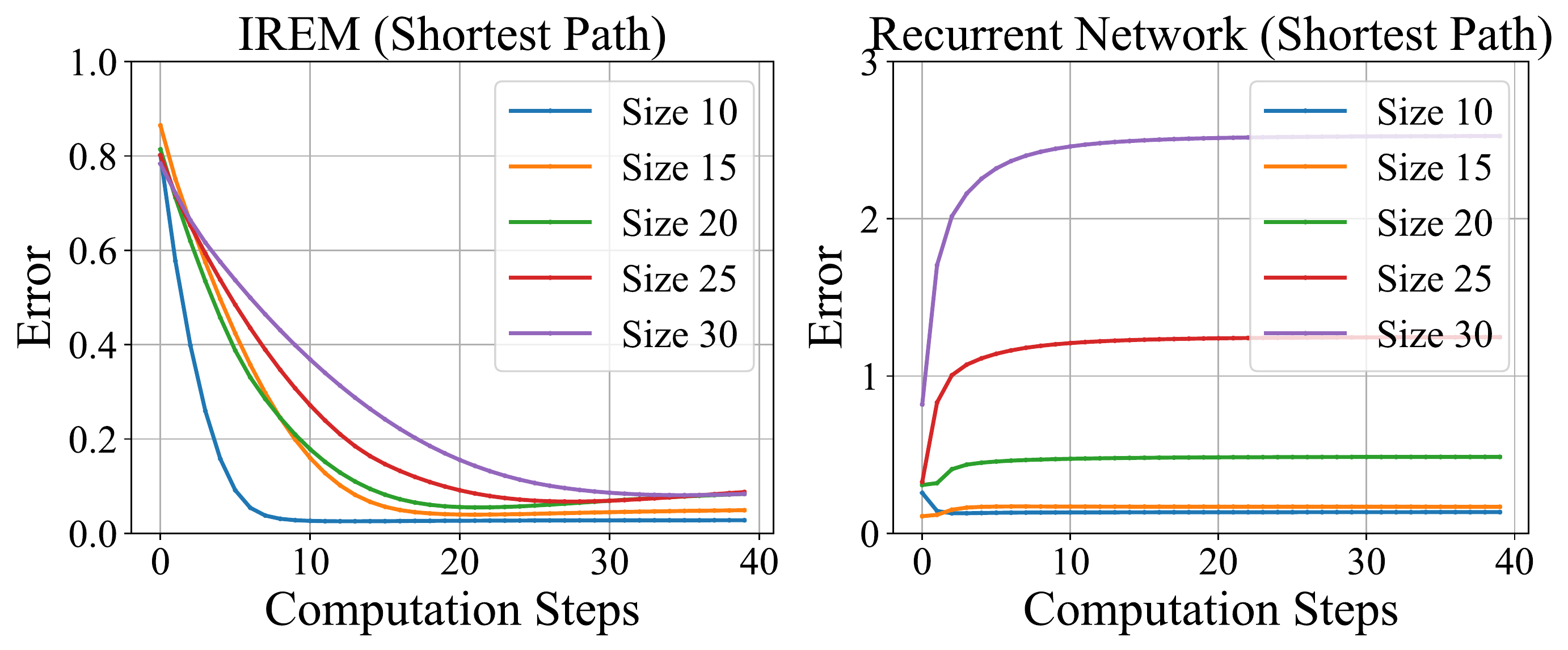}
\vspace{-20pt}
\caption{\small \textbf{Computation Steps vs Problem Size} -- Error on shortest path computation as a factor of graph size and number of computation executed. Models are trained 5 steps of iterative computation on graph sizes of 10. \model is able to generalize computation to larger graphs and a greater number of iterative computation steps, while a recurrent network fails to do so.}
\label{fig:steps_problem}
\vspace{-15pt}
\end{figure}

\myparagraph{Setup.} We first evaluate our approach on graphical algorithmic reasoning. We train models on fully connected graphs with between 2 to 10 nodes, and evaluate performance on larger fully connected graphs with size 15 nodes. We report the underlying elementwise mean squared error between predictions from models and their associated ground truth outputs. We evaluate performance on the three different graphical algorithmic reasoning tasks, aiming to capture different aspects of reasoning, which we detail below, with additional details about each dataset in Appendix \ref{sect:experimental_detail}. 
\begin{enumerate}
    \myitem \textit{Edge Copy}: We first test the ability of models to copy and output the values of all edges in a dense graph that is given as input. This task serves as a simple test for iterative reasoning, and requires a method to sequentially copy over input edge values to the output. 
    \vspace{-10pt}
    \myitem \textit{Connected Components}: Next, we evaluate the ability of models to infer the underlying connected components of a graph. We construct a sparse graph, where 5\% of all fully connected edges exist, and ask models to predict a binary indicator on whether a node of a graph is connected to another for all pairs of nodes in a graph. This task tests structural discovery, an aspect of cognitive reasoning \citep{kemp2008discovery}.
    \myitem \textit{Shortest Path}: Finally, we evaluate the ability of models to compute the shortest path distances between all pairs of nodes in a graph. This task tests for planning, and the underlying calculation necessary to compute the shortest paths between nodes is analogous to that of planning.
\end{enumerate}

\vspace{-5pt}

\myparagraph{Quantitative Results.}  We present quantitative results on graphical algorithmic reasoning in the left column of \tbl{tbl:tbl_algorithm}. During test time, we evaluate on problems with either similar or larger sizes than those seen during training. Across all three tasks, \model outperforms all the compared baselines. This difference in performance is magnified when evaluating on larger graphs during testing. Approaches other than \model fail even on the relatively simple edge copy task. This is due to the fact that the small size of graph networks (hidden size 128), requires methods to iteratively copy different subsets of input edges to output predictions. While \model successfully learns this iterative computation, the compared baselines were unable to do so.

\begin{table}[t]
    \centering

    \scalebox{0.85}{
    \begin{tabular}{ccccc}
        \toprule
        \textbf{Replay}  & \textbf{Truncate} & \textbf{Test} & \textbf{Training} & \textbf{Memory} \\
         \textbf{Buffer} & \textbf{Gradient} & \textbf{Performance} & \textbf{Speed} & \textbf{Usage} \\
        \midrule
        
        No & No & 0.0491  & 87\% & 295\% \\
        Yes & No & 0.0287 & 83\% & 295\% \\
        Yes & Yes & 0.0274  & 100\% & 100\% \\
        \bottomrule
    \end{tabular}
    }
  
    \caption{\small \textbf{Ablations --} Ablations of proposed components of \model on test performance on the shortest path algorithmic reasoning task. The use a replay buffer boosts the underlying performance of \model and truncating gradient backpropogation to the last step of optimization reduces both training time and memory cost. }
     \label{tbl:ablation}
    \vspace{-10pt}
\end{table}

\myparagraph{Adaptive Computation.} Next, we analyze the ability of \model and baselines to adapt its underlying computational budget to different larger problem instances. In \fig{fig:steps_problem}, we illustrate shortest path computation error on different input graphs sizes at test time. While all methods are only trained with 5 steps of iterative computation on size 10 graphs, we find that \model can generalize iterative computation to size 30 graphs. In contrast, the learned iterative baselines, such as the recurrent network in \fig{fig:steps_problem}, fail to do so. We find similar behavior across other graphical algorithmic reasoning tasks.

\myparagraph{Ablations.} We run an ablation analysis on the impact of utilizing a replay buffer to train \model, as well as the effect of truncating backpropagation to the last step of optimization. In \tbl{tbl:ablation}, we find that utilizing a replay buffer significantly improves the performance of \model. We further find that truncating backpropagation to the last step of optimization has a limited impact on \model and greatly reduces the overall memory cost of training, as well as slightly improving the training speed of \model.

\subsection{Continuous Algorithmic Reasoning}
\label{sect:continuous}
\begin{figure}[t]
\centering
\includegraphics[width=1\linewidth]{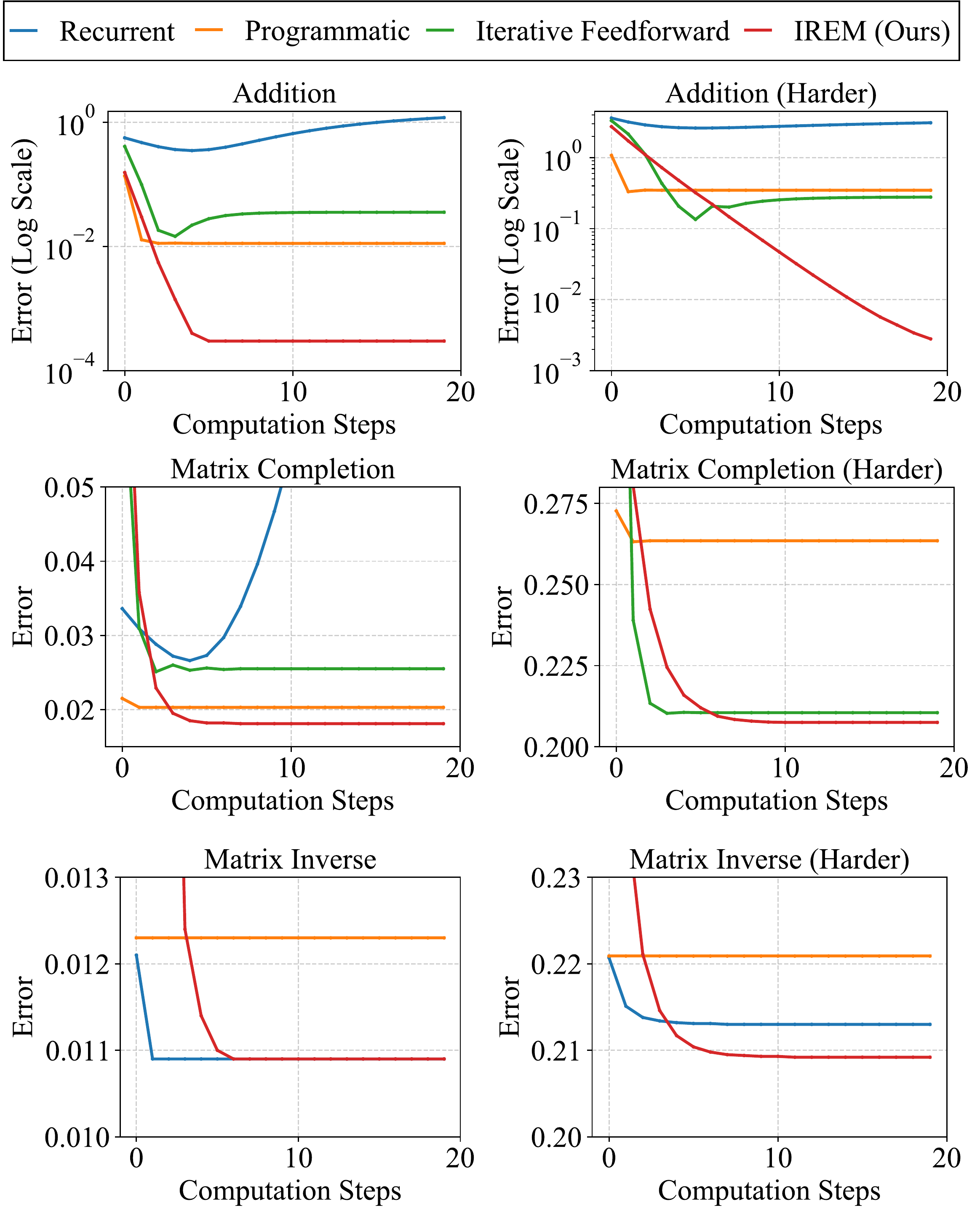}
\vspace{-10pt}
\caption{\small \textbf{Computation Steps vs Problem Difficulty} -- Illustration of MSE error of prediction as a factor of the test time difficulty (harder difficulty right) of the task and computation steps applied. Each model is trained with 5 steps of iterative computation. Models missing in plots have errors greater than the range displayed in the plot. The error of \model improves with the number of underlying algorithmic computation steps, with larger number of computation steps benefiting performance on harder algorithmic tasks. }
\label{fig:continuous_iterative_comp}
\vspace{-5pt}
\end{figure}

\myparagraph{Data Setup.} We next evaluate \model and baselines on continuous algorithmic reasoning tasks. We apply algorithmic operations on input vectors of size 400 (resized to $20 \times 20$ matrices for matrix operations). We report the underlying MSE error between the predictions and the associated ground truth outputs on test problem instances. We evaluate different methods on the following three tasks, aiming to capture different aspects of reasoning, with additional details in the Appendix \ref{sect:experimental_detail}. 

\begin{enumerate}
    \myitem \textit{Addition}: We first evaluate the algorithmic computation of addition. We train networks to add entries in two separate input vectors (element-wise). We construct harder variants of the addition problems at test time by feeding input vectors with larger magnitudes. This task aims to test simple arithmetic reasoning.
    \myitem \textit{Matrix Completion}: Next, we evaluate the algorithmic computation of matrix completion. We mask out $50\%$ of the entries of a low-rank input matrix constructed from two separate low-rank matrices $U$ and $V$, and train networks to reconstruct the original input matrix. We construct harder variants of the matrix completion problem at test time by increasing the complexity of $U$ and $V$. This task aims to test both structural and analogical reasoning, with both shown to be equivalent to matrix completion \citep{lampinen2017analogies}.
    \myitem \textit{Matrix Inverse}: Finally, we evaluate the algorithmic computation of matrix inverse. We train networks to compute the matrix inverse of an input matrix. We construct harder matrix inverse problems by considering less well-conditioned input matrices. This task aims to test the numerical reasoning, with matrix inversion a crucial operation across various numerical algorithms.
\end{enumerate}

\begin{figure}
\centering
\includegraphics[width=1\linewidth]{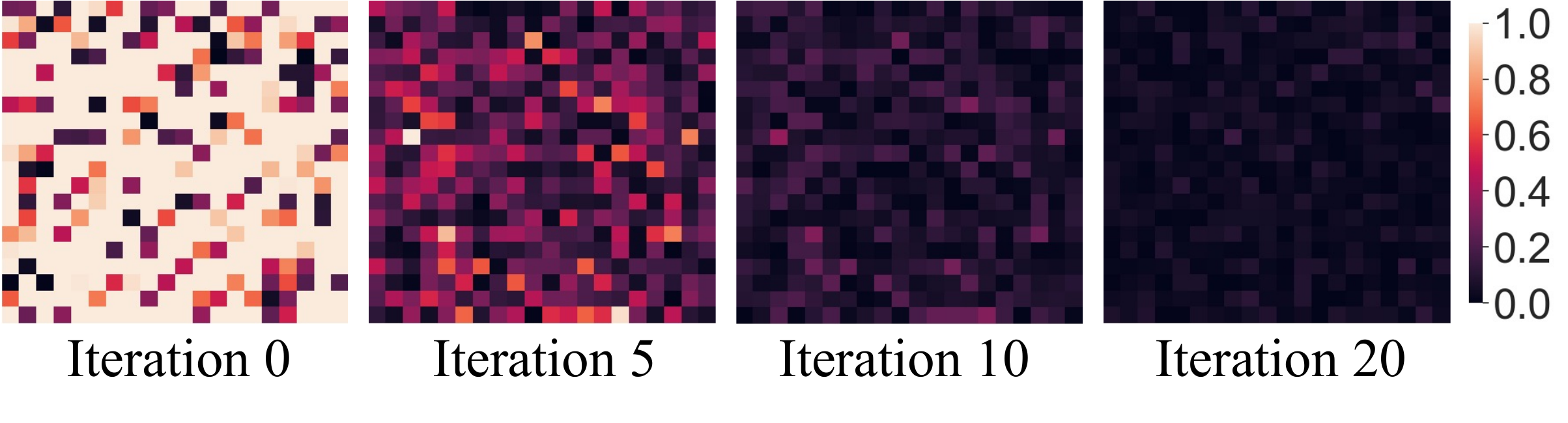}
\vspace{-20pt}
\caption{\small \textbf{Qualitative Illustration of Addition} -- Illustration of per element error on the addition task as a function of underlying number iterative computation steps run with \model. Individual inputs gradually approach ground truth values, with different elements approaching zero error at different rates.} 
\label{fig:qual_optimize}
\vspace{5pt}
\end{figure}

\begin{figure}[t]
\vspace{-5pt}
\includegraphics[width=\linewidth]{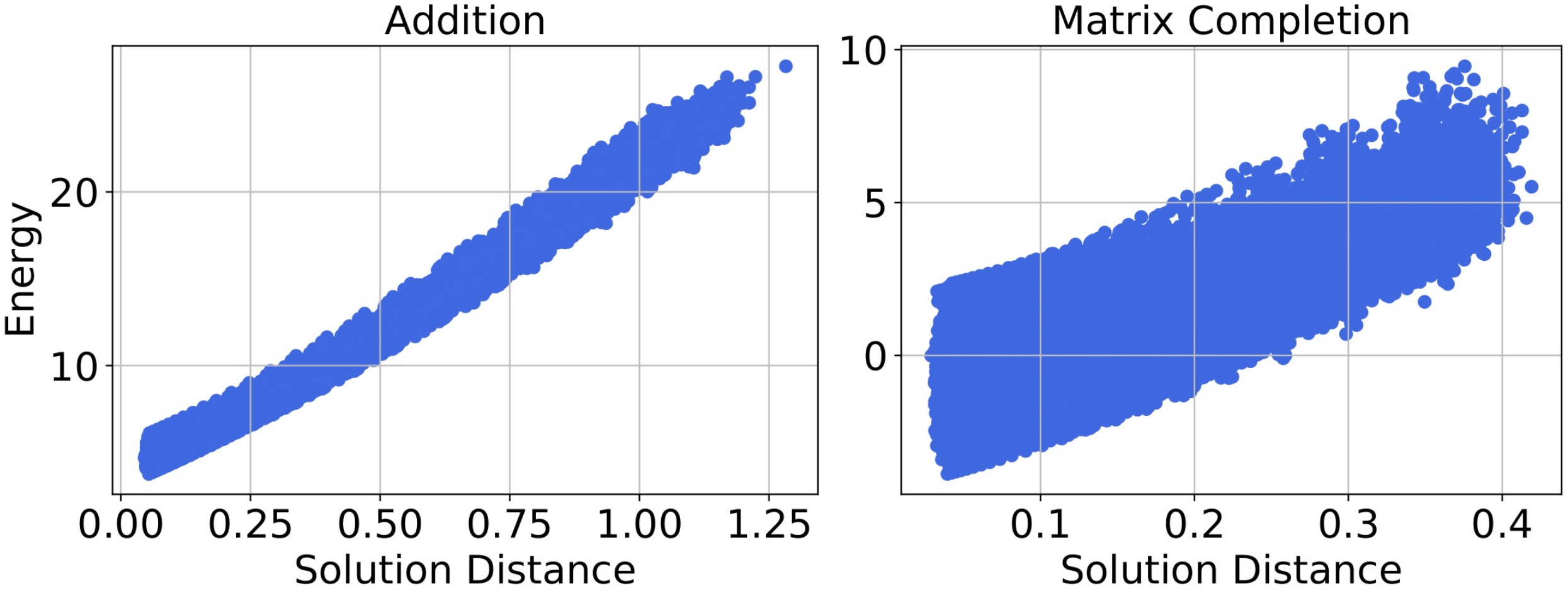}
\vspace{-10pt}
\caption{\small \textbf{Energy Landscape} -- Plot of predicted energy values for $\vy$ and the corresponding MSE distance of $\vy$ from the problem solution. The predicted energy of $\vy$ correlates well with the distance of $\vy$ from the ground truth. Plot for matrix inverse is similar to matrix completion. }
\label{fig:energy_landscape}
\vspace{-5pt}
\end{figure}

\begin{figure*}[t]
\centering
\setlength\arraycolsep{2pt}
\scalebox{0.78}{
$\begin{bmatrix}
  -0.695 & -0.840 & 0.691 \\ 
  0.488 & -0.961 & -0.048 \\
  -0.824 & -0.235 &  0.162
\end{bmatrix} +
\begin{bmatrix}
 -0.590 & 0.972 & -0.544 \\ 
 -0.318 & 0.748 & 0.638 \\
  -0.216 & -0.118 &  0.864
\end{bmatrix} +
\begin{bmatrix}
 -0.676 & -0.688 & 0.422 \\ 
 0.075 & 0.172 & -0.963 \\
  0.698 & 0.837 &  -0.735
\end{bmatrix}  +
\begin{bmatrix}
 0.945 & -0.432 & -0.940 \\ 
 -0.567 & 0.989 & -0.302 \\
  -0.526 & 0.492 &  0.574
\end{bmatrix} 
=
 {\color{blue}\begin{bmatrix}
 -1.033 & -1.076 & -0.433 \\ 
 -0.319 & 1.010 & -0.677 \\
  -0.867 & 0.984 &  0.843
\end{bmatrix} }$

}

\caption{\small \textbf{Addition Composition} -- Illustration of predictions from \model (in blue) when three separate addition executions are nested together. \model is applied on vectors with 400 entries -- we visualize the first 9 elements of inputs and predictions.}
\label{fig:qual_addition_composition}
\vspace{-10pt}
\end{figure*}

\myparagraph{Quantitative Results.} We present quantitative results on each of our three continuous algorithmic reasoning problems on the right side of \tbl{tbl:tbl_algorithm} on test problems with either the same or harder difficulty than training problems. Across all three tasks, we find that \model outperforms baselines, with the underlying difference magnified on harder, out-of-distribution test problems. In particular, on  the task of addition, we find that \model is able to nearly perfectly solve the underlying task even on harder, out-of-distribution test problems. In contrast, all other evaluated iterative and feedforward baselines perform poorly effectively solve the underlying problem. A primary difficulty is that underlying input vectors, both of size 400, are large compared to the underlying hidden unit size of 512, thus requiring the networks to iteratively reason and execute the algorithmic computation on subsets of the input. 

\begin{table}
\small\setlength{\tabcolsep}{5.5pt}
\centering
\begin{tabular}{lcc}
\toprule
      {\bf Step Size} & {\bf Same Difficulty} & {\bf Harder Difficulty} \\
      \midrule
      10 & 0.0003 & 0.0021 \\
      30 & 0.0003 & 0.0020 \\
      100 & 0.0003 & 0.0021 \\
      300 & 0.0004 & 0.0023 \\
      1000 & 0.0004 & 0.0025 \\
    \bottomrule
\end{tabular}
\caption{\small \textbf{Ablation Analysis of Step Size in \model--} Analysis of training step size of \model  performance on the continuous addition reasoning task.}
\label{tbl:ablation_step_size}
\vspace{-20pt}
\end{table}

\myparagraph{Qualitative Visualization.} Next, we visualize the underlying iterative computation learned by \model and baselines. In \fig{fig:continuous_iterative_comp}, we illustrate the prediction error of different methods as a function of the number of computation steps applied. As we increase the number of iterative computation steps, the underlying performance of \model continues to improve. In particular, iterative computation helps more substantially on harder variants of the algorithmic problem (right column), such as addition. In contrast, several iterative methods show significant degradation of performance with increased iterative computation.

We further qualitatively visualize the underlying iterative computation learned by \model. In \fig{fig:qual_optimize}, we visualize the element-wise mean square error of the predicted solution as a function of the number of iterative reasoning steps applied on the addition task.  We find that energy minimization gradually refines a predicted solution to the ground truth additive answer, with different elements of the solution exhibiting different convergence rates.

\myparagraph{Energy Landscape.} \model parameterizes an energy landscape across all possible solutions for a given problem. Such an energy landscape enables us to assess the relative quality of solutions dependent on their associated energies, and further gives a natural objective to terminate iterative computation when an underlying local energy minimum is reached. In \fig{fig:energy_landscape}, we visualize the predicted energy of different candidate solutions and their corresponding MSE distances from the ground truth answer. We find that across different continuous algorithmic tasks, the underlying energy value assigned to a candidate solution is well correlated with its distance from the ground truth answer, with low energy solutions close to the ground truth.

\myparagraph{Sensitivity to Step Size.} We assess the performance of \model under different values of step size $\lambda$ using during energy optimization at training. We consider the continuous addition task in \tbl{tbl:ablation_step_size}. We find that the underlying performance is not sensitive to hyperparameter choice for step size, and utilize a fixed step size of 100 across our experiments.

\subsection{Recursive Algorithmic Computation}

\myparagraph{Setup.} We further evaluate the ability of algorithms represented by \model to be recursively applied on inputs. Recursively nesting algorithms enable complex computations, but require learned networks to be robust to out-of-distribution outputs from prior algorithmic execution.

\begin{table}
\small
\setlength{\tabcolsep}{3.5pt}
\centering
\scalebox{0.9}{
\begin{tabular}{lc@{}c@{}c}
    \toprule
      &  & {\bf Composed Operations} \\
      \cmidrule(lr){2-4} 
      {\bf Method} & 2 & 5 &  10 \\
      \midrule
     Feedforward & 0.0445 & 0.2717  &  0.8898 \\
      Recurrent &  2.1377  & 3.1861 & 4.8706 \\
       Programmatic & 0.0203 & 0.1068 & 0.4587 \\
       Iterative Feedforward & 0.0826 &  0.5930 & 3.6004  \\
       \model (Ours) & \textbf{0.0014} & \textbf{0.0078} & \textbf{0.0422} \\
    \bottomrule
\end{tabular}
}

\caption{\small \textbf{Algorithmic Composition --} Test performance when composing multiple instances of the addition operation.  Error is reported using element-wise mean square error. \model is able to generalize well when composing algorithmic computation.}
\label{tbl:tbl_compose}
\end{table}
\begin{figure}[t]
\begin{center}
\vspace{-5pt}
\includegraphics[width=0.8\linewidth]{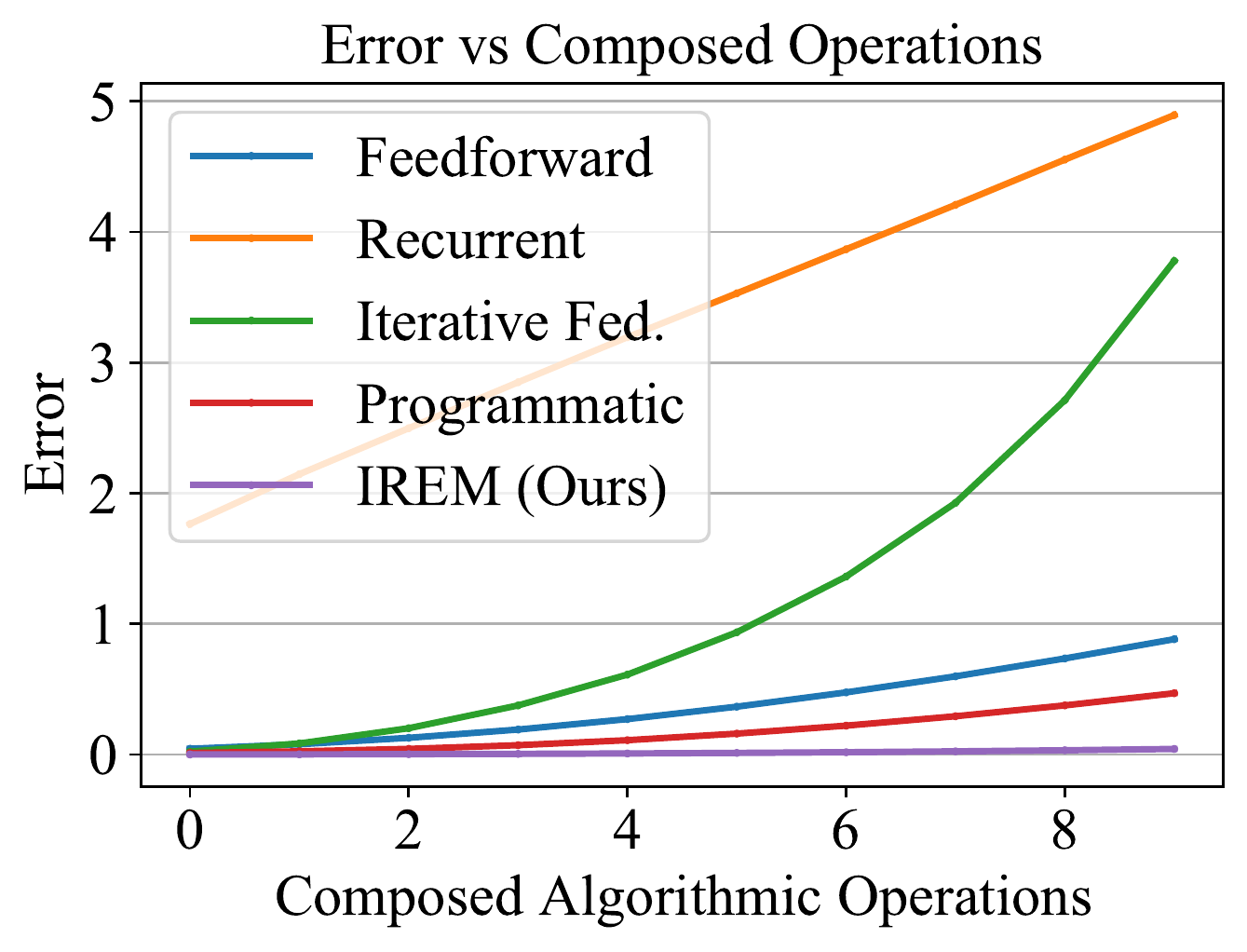}
\vspace{-10pt}
\caption{\small \textbf{Error Composing Algorithmic Computation} -- Plot of error of predictions as factor of the number of composed algorithmic operations. \model exhibits lower error when nesting a series of algorithmic operations together.}
\label{fig:comp_plot}
\end{center}

\vspace{-20pt}
\end{figure}

We consider recursive applications of a learned  algorithmic operator $\text{Alg}_\theta(\cdot)$ representing addition as introduced in \sect{sect:continuous}. We evaluate the element-wise mean square error of the predicted output $\hat{y}$ of recursively applying the learned algorithmic operator $k$ times
\begin{equation}
    \hat{y}^k = \text{Alg}_\theta(\hat{y}^{k-1}, y_k),
\end{equation}
with the corresponding ground truth solution being $\sum_k y_k$, for different values $k$ of recursive application.

\myparagraph{Quantitative Results.} We report the results of recursively applying each learned algorithmic operator between two to ten times in \tbl{tbl:tbl_compose}. We find that \model supports the most stable recursive application of algorithmic operators. \model exhibits significantly lower error than all compared baseline, due to its ability to utilize iterative computation to deal with out-of-distribution inputs and to accurately compute intermediate algorithmic outputs.

\myparagraph{Qualitative Results.} We illustrate error as a factor of the number of applied algorithmic operations in \fig{fig:comp_plot}, and find that the error of predictions from \model rises slowly in comparison to other baselines. We further visualize the nested algorithmic predictions from \model. We illustrate the inferred array sum predicted by \model when four separate inputs are summed in \fig{fig:qual_addition_composition}. As seen above, our approach enables us to closely approximate the addition of four input matrices (with the first nine entries shown).

\section{Conclusion and Limitations}

In this paper, we present \model, a new approach towards iterative computation, by formulating it as an energy minimization process. We illustrate, on both continuous and graphical domains, how iterative computation utilizing \model enables better algorithmic performance, as well as generalization to more complex instances of problems. We further illustrate how the underlying algorithmic computation learned by \model may be nested to implement more complex algorithmic computations.

Iterative reasoning with \model has several limitations. First, while \model substantially outperforms existing iterative methods on tasks where output solutions have high dimensionality, limited gains are obtained when \model is executed on problems with lower dimensionality solutions (such as parity prediction).  Second, as training and inferring solutions with \model relies on continuous gradient optimization, \model struggles when output solutions have discrete values. An interesting line of future work would be to explore how discrete optimization could be integrated with training \model to solve such discrete valued problems. Finally, since the training procedure of \model requires backpropagation across gradient optimization steps, it is computationally expensive. An interesting line of future work could be exploring alternative ways to train an energy function for reasoning, such as utilizing gradient-free optimization.

\section{Acknowledgements}

We thank Ben Poole for feedback and helpful comments on a early version of the manuscript. Yilun Du is supported by a fellowship from the National Science Foundation.

\bibliography{references}
\bibliographystyle{icml2022}

\clearpage
\textbf{\huge{Appendix}}
\appendix

\appendix
In this appendix we provide additional evaluation and details on \model. First, we illustrate how \model may be applied to an existing image iterative reasoning task in Appendix~\ref{sect:image_denoise}. Next we discuss how we may utilize optimization to execute \model with an external scratchpad in Appendix~\ref{sect:scratchpad}.  We further discuss additional experimental details on our evaluated algorithmic tasks in Appendix~\ref{sect:experimental_detail}. Finally, we discuss individual model architectures used in Appendix~\ref{sect:model_architecture}

\section{Image Denoising}
\label{sect:image_denoise}

\myparagraph{Setup} We compare \model with existing approaches on an existing image based iterative reasoning benchmark from \citep{chen2020learning}. The benchmark task is to denoise images with various levels of Gaussian noise corruption added. Harder denoising tasks are constructed at test time by adding larger amounts of noise to input images. We directly compare with baselines and numbers for UNLNet, DnCNN and DNCNN-stop from \citep{chen2020learning} and utilize the authors provided training code to train \model with five steps of iterative reasoning. 

\myparagraph{Quantitative Results} We quantitatively evaluate the performance of \model and baselines in terms of PSNR (numbers for baselines directly from ~\citep{chen2020learning}) in \tbl{tbl:image_denoise}. While we found that \model performed similarly to existing approaches when the test noise corruption is similar to that of training, \model significantly outperformed the compared baselines when evaluated on harder images at test time which exhibited substantially larger amounts of noise corruption then seen during training. We found that such performance gain was due to iterative computation executed by \model.  While we trained \model with 5 steps of reasoning, at test time we found that running up to 30 steps of reasoning at noise level $\sigma=65$ improved performance, while running up to 100 steps of reasoning at noise level $\sigma=75$ further improved performance.

\begin{figure}[H]
\centering
\includegraphics[width=0.95\linewidth]{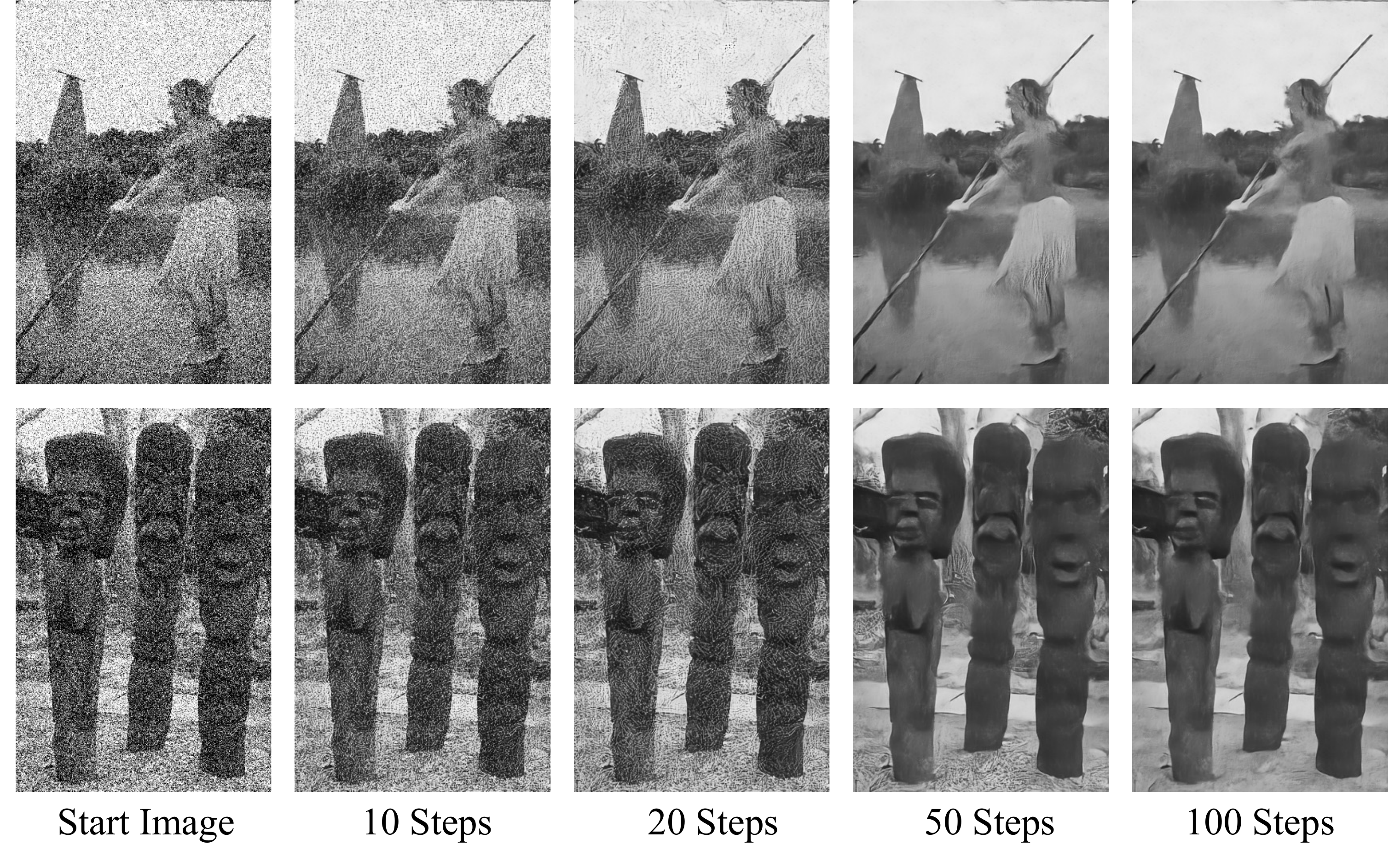}
\vspace{-10pt}
\caption{\small \textbf{Qualitative Illustration of Iterative Image Denoising} -- Illustration of iterative denoising using \model. Images continue to be cleaner after up to 100 iterative steps of computation.} 
\label{fig:image_denoise}
\end{figure}

\myparagraph{Qualitative Results} We qualitatively illustrate \model being applied to an image corrupted with an unseen noise level $\sigma=75$ (significantly larger than what is seen during training) in \fig{fig:image_denoise}. As seen in \fig{fig:image_denoise}, images continue to become clearer, even after 50 steps of iterative computation using \model (with dress in the top row and floor in the bottom row becoming clearer).

\section{Iterative Reasoning with a Scratchpad}
\label{sect:scratchpad}

We next discuss how we may incorporate an external computational scratchpad when executing an iterative computation with \model. To enable the processing of an external scratchpad, $\vz \in \mathbb{R}^D$, we construct a EBM, $E_\theta(\vx, \vy, \vz): \mathbb{R}^N \times \mathbb{R}^M  \times \mathbb{R}^D \rightarrow \mathbb{R}$, which takes as input an input problem $\vx$, a candidate solution $\vy$, and a scratchpad state $\vz$.

\begin{table}[t]
    \centering

    \scalebox{0.8}{
    \begin{tabular}{lcccc}
        \toprule
        $\bm{\sigma}$ &  \textbf{UNLNet} & \textbf{DnCNN} & \textbf{DnCNN-stop} & \textbf{\model (Ours)} \\
        \midrule
        \textbf{45} & 26.48 & \textbf{26.56} & 26.48 & 26.52 \\ 
        \textbf{55} & 25.64 & 25.71 & 25.79 & \textbf{25.85} \\
        \midrule
        \textbf{*65} & - & 22.19 & 23.56 & \textbf{23.87} \\
        \textbf{*75} & - & 17.90 & 18.62 & \textbf{23.43}\\
        \bottomrule
    \end{tabular}
    }

    \caption{\small \textbf{Image Denoising --} Reconstruction PSNR results for \model and baseline comparisons. The $*$ indicates noise levels of 65 and 75 which are not in the training set. Numbers for baselines directly reported from \citep{chen2020learning}. \model generalizes substantially better to more complex denoising tasks. }
     \label{tbl:image_denoise}
    \vspace{-10pt}
\end{table}
We then define our predicted solution 
\begin{equation}
    \hat{\vy} = \argmin_{\vy} \min_{\vz} E_\theta(\vx, \vy, \vz),
\end{equation}
where optimize over both output solutions $\vy$ and an external scratchpad $\vz$ at prediction time. Analogous to the training procedure in the main paper, we may obtain a fast approximation of $\hat{\vy}$ as $\vy^N$ training time by jointly optimizing $\vy$ and $\vz$ 
\begin{equation}
    \begin{aligned}
    \vy_i^n &= \vy_i^{n-1} - \lambda \nabla_{\vy} E_{\theta} (\vx_i, \vy_i^{n-1}, \vz_i^{n-1}) \\
    \vz_i^n &= \vz_i^{n-1} - \lambda \nabla_{\vz} E_{\theta} (\vx_i, \vy_i^{n-1}, \vz_i^{n-1}).
    \end{aligned}
    \label{eqn:opt_mult_mem}
\end{equation}
We then analogously minimize the corresponding loss:
\begin{equation}
      \mathcal{L}_{\text{MSE}}(\theta) = \| \vy_i^N - \vy_i \|^2,
\end{equation}

We provide the overall pseudocode for training \model with external memory in Algorithm ~\ref{alg:train_mem}.

\begin{algorithm}
\small
\begin{algorithmic}
    \STATE \textbf{Input:} Data Dist $p_D(\vx, \vy)$, Replay Buffer $\mathcal{B}$, Step Size $\lambda$, Number of Steps $N$, EBM $E_\theta(\cdot)$, Uniform Distribution $U(-1, 1)$
    \STATE $\mathcal{B} \gets \varnothing$
    \WHILE{not converged}
    
    \STATE \emph{$\triangleright$ Sample data and candidate solutions from $p_d$ and replay buffer $\mathcal{B}$}
    \STATE $\vx_i, \vy_i \sim p_D$, $\tilde{\vy}_i^0, \tilde{\vz}_i^0 \sim \mathcal{U}(-1, 1)$
    \STATE $\vx_i^b, \vy_i^b, \tilde{\vy}_i^{b}, \tilde{\vz}_i^{b} \sim B$
    \STATE $\vx_i, \vy_i, \tilde{\vy}_i^0, \tilde{\vz}_i^0 \gets \vx_i \cup \vx_i^b, \vy_i \cup \vy_i^b, \tilde{\vy}_i^0 \cup \tilde{\vy}_i^b, \tilde{\vz}_i^0 \cup \tilde{\vz}_i^b$
    \vspace{2mm}
    
    \STATE \emph{$\triangleright$ Generate low energy solutions through optimization:}
    \FOR{sample step $n = 1$ to $N$}
    \STATE $\tilde{\vy}_i^n \gets \tilde{\vy}_i^{n-1} -  \lambda \nabla_\vy E_\theta (\vx_i, \tilde{\vy}_i^{n-1}, \tilde{\vz}_i^{n-1})$
    \STATE $\tilde{\vz}_i^n \gets \tilde{\vz}_i^{n-1} -  \lambda \nabla_\vz E_\theta (\vx_i, \tilde{\vy}_i^{n-1}, \tilde{\vz}_i^{n-1})$
    \ENDFOR 
    \vspace{2mm}
    
     \STATE \emph{$\triangleright$ Optimize objective $\mathcal{L}_{\text{MSE}}$ wrt $\theta$:} 
    \STATE $\Delta \theta \gets \nabla_\theta \sum_{n=1}^N \|\tilde{\vy}^N_i - \vy_i\|^2 $
    \STATE Update $\theta$ based on $\Delta \theta$ using Adam optimizer 
    
    \vspace{2mm}
    \STATE \emph{$\triangleright$ Update replay buffer $\mathcal{B}$}
    \STATE $\mathcal{B} \gets \mathcal{B} \cup (\vx_i, \vy_i, \tilde{\vy}_i^{N}, \tilde{\vz}_i^N)$
    \ENDWHILE

  \end{algorithmic}
 \caption{\model Training with External Memory}
 \label{alg:train_mem}
 \end{algorithm}

\section{Experimental Details}
\label{sect:experimental_detail}

\myparagraph{Graphical Algorithmic Computation}  Models were trained in approximately 2 hours on a single Nvidia Titan X GPU using a training batch size of 64 and the Adam optimizer with learning rate 1e-4. Each model was trained for 10,000 iterations and evaluated on 1000 test problems. Each model was trained with five steps of iterative computation, with PonderNet trained with a halting geometric distribution of 0.8. Below, we provide additional numerical details about each of the evaluated algorithmic tasks.

\begin{enumerate}
    \myitem \textit{Edge Copy:} We randomly sample a value for each edge in a fully connected graph with a uniform value between -1 and 1. Models are then tasked with replicating the value of each individual edge in the graph in the final output prediction. 
    \item \textit{Connected Components:} We randomly zero-out 95\% of the edges in a fully connected graph. Models are then tasked with predicting the pairwise connectively of all possible pairs of nodes in the graph.
    \item \textit{Shortest Path:} We randomly sample an edge distance between 0 and 1 for each edge in a fully connected graph. Models are then tasked with predicting the pairwise shortest distance between all possible pairs of nodes in the graph.
\end{enumerate}

\myparagraph{Continuous Algorithmic Computation} Models were trained in approximately 2 hours on a single Nvidia Titan X GPU using a training batch size of 128 and the Adam optimizer with learning rate 1e-4. Each model was trained for 10,000 iterations and evaluated on 1000 test problems. Each model was trained with five steps of iterative computation, with PonderNet trained with a halting geometric distribution of 0.8. We further provide individual dataset details below.

\begin{enumerate}
    \item \textit{Addition: } We randomly construct two separate vectors, each with 400 elements, with each element in the vector randomly sampled between -1 and 1. Models are then tasked with summing up the elements in each vector element-wise. When constructing more difficult addition problems at test time, each element in the vector is randomly sampled between -2.5 and 2.5.
    \item \textit{Matrix Completion:} We randomly construct a low-rank matrix $M$ represented as the $M = U^{T}V + 0.1\mathcal{N}(0, 1)$, where $U$ and $V$ are $10 \times 20$ matrices, with each individual elements in $U$ and $V$ sampled from  $\mathcal{N}(0, 0.22)$ . Models are given 50\% of the entries of $M$ are tasked with recovering all entries of $M$. When constructing more difficult matrix completion problems at test time, each element in $U$ and $V$ are sampled from $
    \mathcal{N}(0, 0.47)$.
    \item \textit{Matrix Inverse:} We randomly construct a well conditioned invertible matrix $M = R + R^T + 0.5 * I$, where $R$ is a random matrix, with individual elements sampled between -1 and 1. Models are tasked with computing the matrix inverse of $M$. When constructing more difficult matrix inversion problems at test time, we make $M$ less well-conditioned by setting $M = R + R^T + 0.1 * I$.
\end{enumerate}

\section{Model Architectures}
\label{sect:model_architecture}

\myparagraph{Graphical Algorithmic Computation} For each iterative and feedforward method, we utilize the GINEConv layer from \citep{hu2019strategies}, where $\text{GINEConv(}128, 128)$ refers to a graph convolution operator with node features 128 and edge feature 128 . An input problem instance $\vx$ consists of a set of nodes features $\vv$ and edge features $\ve$. To parameterize the EBM in \model $E_\theta(\vx, \vy)$ we concatenate the optimized prediction $\hat{\vy}$, with $\ve$, which then utilized in the GINEConv layer. To obtain per edge predictions for baselines, we pairwise concatenate node features for  the given edges,  and apply an FC layer to obtain the corresponding prediction following \citep{zhang2018link}.   We specify the architecture for \model in \tbl{tbl:graph_ebm}, the architecture for feedforward and iterative feedforward baselines in \tbl{tbl:graph_ff}, the architecture for recurrent baselines in \tbl{tbl:graph_recurrent}, and the architecture for programmatic execution baselines in \tbl{tbl:graph_pondernet}. All models have roughly the same number of underlying parameters.

\begin{figure}[H]
\begin{minipage}{0.45\linewidth}
\centering
\small
\begin{tabular}{c}
    \toprule
    GINEConv(128, 128) \\
    \midrule
    GINEConv(128, 128) \\
    \midrule
     GINEConv(128, 128) \\
    \midrule
    Linear 128 $\rightarrow$ 1 \\ 
    \bottomrule
\end{tabular}
\captionof{table}{The model architecture for \model on graphical tasks.}
\label{tbl:graph_ebm}
\end{minipage}%
\hfill
\begin{minipage}{0.45\linewidth}

\centering
\small
\begin{tabular}{c}
    \toprule
    GINEConv(128, 128) \\
    \midrule
    GINEConv(128, 128) \\
    \midrule
     GINEConv(128, 128) \\
    \midrule
    Linear 256 $\rightarrow$ Output Dim \\ 
    \bottomrule
\end{tabular}
\captionof{table}{The model architecture for feedforward and iterative feedforward baselines on graphical tasks. }
\label{tbl:graph_ff}
\end{minipage}
\end{figure}

\begin{figure}[H]
\begin{minipage}{0.45\linewidth}
\centering
\small
\begin{tabular}{c}
    \toprule
     GINEConv(128, 128) \\
    \midrule
    LSTM(128) \\
    \midrule
     GINEConv(128, 128) \\
    \midrule
    Linear 256 $\rightarrow$ Output Dim \\ 
    \bottomrule
\end{tabular}
\captionof{table}{The model architecture for recurrent baseline on graphical tasks.}
\label{tbl:graph_recurrent}
\end{minipage}%
\hfill
\begin{minipage}{0.45\linewidth}

\centering
\small
\begin{tabular}{c}
    \toprule
     GINEConv(128, 128) \\
    \midrule
    GINEConv(128) \\
    \midrule
     GINEConv(128, 128) \\
    \midrule
    Linear 256 $\rightarrow$ Output Dim\\ 
    \midrule
    Linear $\rightarrow$ 1 \\
    \bottomrule
\end{tabular}
\captionof{table}{The model architecture for PonderNet baseline on graphical tasks.}
\label{tbl:graph_pondernet}
\end{minipage}
\end{figure}

\myparagraph{Continuous Algorithmic Computation} For each iterative and feedforward method, we utilize a MLP to implement continuous algorithmic computation. To parameterize the EBM in \model $E_\theta(\vx, \vy)$, we concatenate $\vx$ and $\vy$ together as input into the network. We utilize the ReLU activation in all networks except \model, where we utilize the Swish activation.  We specify the architecture for \model in \tbl{tbl:continuous_ebm}, the architecture for feedforward and iterative feedforward baselines in \tbl{tbl:continuous_ff}, the architecture for recurrent baselines in \tbl{tbl:continuous_recurrent}, and the architecture for programmatic execution baselines in \tbl{tbl:continuous_pondernet}. All models have roughly the same number of underlying parameters.

\begin{figure}[H]
\begin{minipage}{0.45\linewidth}
\centering
\small
\begin{tabular}{c}
    \toprule
    Linear 512 \\
    \midrule
    Linear 512 \\
    \midrule
    Linear 512 \\
    \midrule
    Linear $\rightarrow$ 1 \\ 
    \bottomrule
\end{tabular}
\captionof{table}{The model architecture for \model on continuous tasks.}
\label{tbl:continuous_ebm}
\end{minipage}%
\hfill
\begin{minipage}{0.45\linewidth}

\centering
\small
\begin{tabular}{c}
    \toprule
    Linear 512 \\
    \midrule
    Linear 512 \\
    \midrule
    Linear 512 \\
    \midrule
    Linear $\rightarrow$ Output Dim \\ 
    \bottomrule
\end{tabular}
\captionof{table}{The model architecture for feedforward and iterative feedforward baselines on continuous tasks.}
\label{tbl:continuous_ff}
\end{minipage}
\end{figure}

\begin{figure}[H]
\begin{minipage}{0.45\linewidth}
\centering
\small
\begin{tabular}{c}
    \toprule
    Linear 196 \\
    \midrule
    LSTM 196\\
    \midrule
    Linear $\rightarrow$ Output Dim\\
    \bottomrule
\end{tabular}
\captionof{table}{The model architecture for recurrent baseline on continuous tasks.}
\label{tbl:continuous_recurrent}
\end{minipage}%
\hfill
\begin{minipage}{0.45\linewidth}

\centering
\small
\begin{tabular}{c}
    \toprule
    Linear 512 \\
    \midrule
    Linear 512 \\
    \midrule
    Linear 512 \\
    \midrule
    Linear $\rightarrow$ Output Dim\\ 
    \midrule
    Linear $\rightarrow$ 1 \\
    \bottomrule
\end{tabular}
\captionof{table}{The model architecture for PonderNet baseline on continuous tasks.}
\label{tbl:continuous_pondernet}
\end{minipage}
\end{figure}

\end{document}